%% file: main.tex
\documentclass[letterpaper, 10 pt, onecolume, conference]{ieeeconf}  % Comment this line out if you need a4paper

\IEEEoverridecommandlockouts                              % This command is only needed if 

\usepackage{amsmath} % assumes amsmath package instaThread Reconstr OUTLINE 2 RAL (Copy)lled
\usepackage{amssymb}  % assumes amsmath package installed
\usepackage{graphicx}
\usepackage{balance}
\usepackage{lipsum}
\usepackage{algpseudocode}
\usepackage{multirow}
\usepackage{dsfont}
\usepackage{booktabs}
\usepackage{xcolor}
\usepackage{adjustbox}
\usepackage{stmaryrd}
\usepackage{graphicx} % for pdf, bitmapped graphics files
\usepackage{multirow}
\usepackage{nicefrac, xfrac}
\usepackage{amsmath} % assumes amsmath package installed
\usepackage{amssymb}  % assumes amsmath package installed
\usepackage{xcolor}
\usepackage{lipsum}
\usepackage{booktabs}
\usepackage{hyperref}
\usepackage{cleveref}
\usepackage{adjustbox}
\usepackage{mathtools}
\usepackage{float}
\usepackage{amssymb}% http://ctan.org/pkg/amssymb
\usepackage{pifont}

\usepackage[ruled,vlined,linesnumbered]{algorithm2e}
\definecolor{ForestGreen}{rgb}{0.133, 0.545, 0.133}

\SetCommentSty{mycommfont}

\definecolor{feicolor}{RGB}{201, 104, 104}

\title{\LARGE
\textbf{AutoPeel}: Adhesion-aware Safe Peeling Trajectory Optimization for Robotic Wound Care
}

\author{Xiao Liang$^{*1}$, Youcheng Zhang$^{*1}$, Fei Liu$^2$, Florian Richter$^1$, Michael Yip$^1$ % <-this % stops a space
\thanks{$^*$ \textit{These authors contributed equally to this work}}
\thanks{This project was funded by the US Army Telemedicine and Advanced Technologies Research Center (TATRC), NIH \#1R01CA278703-01, and NSF Career Award \#2045803}% <-this % stops a space
\thanks{$^{1}$ Department of Electrical and Computer Engineering, University of California San Diego, La Jolla, CA 92093, USA {\tt\small \{x5liang, yoz024, frichter, yip\}@ucsd.edu}}%
\thanks{$^{2}$ Department of Electrical Engineering \& Computer Science, University of Tennessee, Knoxville, TN 37996, USA {\tt\small \{fliu33\}@utk.edu}}%
}

\begin{document}
\newcommand{\statew}{\mathbf{x}_t}
\newcommand{\states}{\mathbf{x}^{s}_t}
\newcommand{\stated}{\mathbf{x}^{d}_t}
\newcommand{\statedad}{\mathbf{\bar x}^{d}_t}
\newcommand{\graph}{\mathcal{G}}
\newcommand{\graphd}{\mathcal{G}^d}
\newcommand{\graphs}{\mathcal{G}^s}
\newcommand{\graphsd}{\mathcal{G}^{d\shortrightarrow s}}
\newcommand{\argmin}{\mathop{\mathrm{argmin}}} 
\newcommand{\argmax}{\mathop{\mathrm{argmax}}} 
\newcommand{\vre}{V^{\text{det}}}

\renewcommand*{\figureautorefname}{Fig.}
\renewcommand*{\subsectionautorefname}{Sec}
\renewcommand*{\sectionautorefname}{Sec}
\renewcommand*{\equationautorefname}{Eqn.}
\renewcommand*{\tableautorefname}{Tab.}

\maketitle
\thispagestyle{empty}
\pagestyle{empty}

%%%%%%%%%%%%%%%%%%%%%%%%%%%%%%%%%%%%%%%%%%%%%%%%%%%%%%%%%%%%%%%%%%%%%%%%%%%%%%%%
\begin{abstract}
Chronic wounds, including diabetic ulcers, pressure ulcers, and ulcers secondary to venous hypertension, affects more than 6.5 million patients and a yearly cost of more than \$25 billion in the United States alone.
Chronic wound treatment is currently a manual process, and we envision a future where robotics and automation will aid in this treatment to reduce cost and improve patient care.
In this work, we present the development of the first robotic system for wound dressing removal which is reported to be the worst aspect of living with chronic wounds.
Our method leverages differentiable physics-based simulation to perform gradient-based Model Predictive Control (MPC) for optimized trajectory planning.
By integrating fracture mechanics of adhesion, we are able to model the peeling effect inherent to dressing adhesion.
The system is further guided by carefully designed objective functions that promote both efficient and safe control, reducing the risk of tissue damage.
We validated the efficacy of our approach through a series of experiments conducted on both synthetic skin phantoms and real human subjects.
Our results demonstrate the system's ability to achieve precise and safe dressing removal trajectories, offering a promising solution for automating this essential healthcare procedure.
\end{abstract}

\input{tex_files/introduction}
\input{tex_files/related_work}

\input{tex_files/method}

\input{tex_files/experiments}
\input{tex_files/conclusion}

\clearpage
\balance
\bibliographystyle{IEEEtran}
\bibliography{root}

\end{document}

%% file: tex_files/introduction.tex
\section{INTRODUCTION}
Chronic wounds, including diabetic ulcers, pressure ulcers, and ulcers secondary to venous hypertension \cite{weinzweig2010plastic}, are wounds that do not heal in a normal and timely manner \cite{siddiqui2010chronic}.
In the United States, chronic wounds affect more than 6.5 million patients \cite{olsson2019humanistic} and estimated more than \$25 billion is spent annually on treatments \cite{woundcare, han2017chronic}.
This estimate is rapidly expanding due to a rise in healthcare costs, an aging population, and a rise in diabetes and obesity. 
A cost-effective robotic system for wound care could play a vital role in mitigating the cost and improving patient care.

Pioneer research works in robotics and automation have explored various aspects of wound care including wound debridement \cite{schoeb2018robotic} and wound filling \cite{jafari2018robot}.
%, and wound 3D reconstruction \cite{filko20232d, filko2021automatic}. 
While these works explore the nature of wound care at its origination, the numbers pale in comparison to instances of caring for those chronic wounds that do not heal and remain for extended periods of time, or indefinitely. In particular, this paper is focused on specifically wound dressing removal.
Maintaining an environment with low oxygen tension through moist occlusive dressings helps support the inflammatory phase of wounds \cite{ke2006hypoxia, jones2006wound}.
Therefore, caregivers manually change dressings on a regular basis, however, the acute and chronic pain of chronic wounds is amplified during this treatment \cite{price2007managing}.
In fact, the pain from wound dressing changes is reported to be the worst part of living with chronic wounds \cite{price2008dressing}.
% This process, traditionally carried out by caregivers, requires careful attention to avoid damaging fragile tissue or causing unnecessary pain during dressing changes.
While there have been efforts in developing sophisticated dressings for minimizing pain while maintaining moisture for the wound \cite{kirkcaldy2023strategies}, there is no prior work in automating the care-takers role who has a significant impact on wound redressing \cite{brett2006impact}.

\begin{figure}[t]
    \centering
    \includegraphics[width=0.48\textwidth]{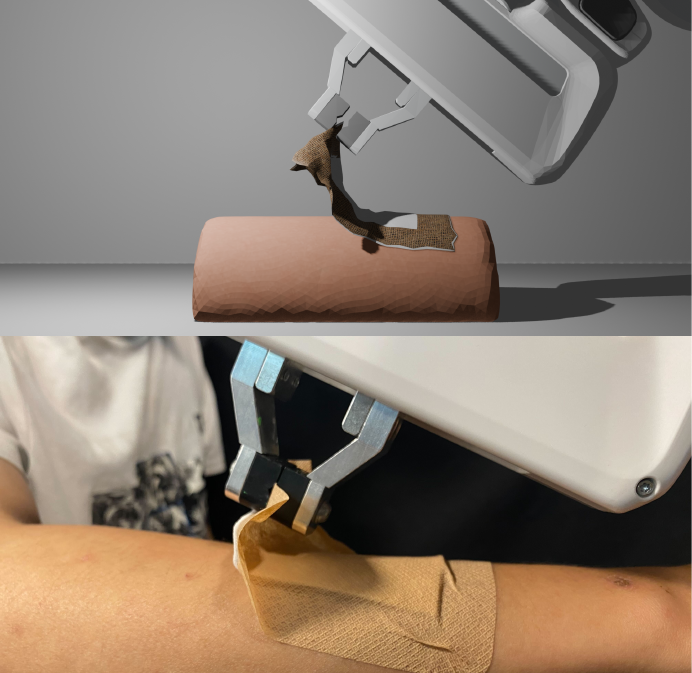}\vspace{-1.5em}
    \caption{An illustration of the robotic wound dressing change concept presented in this paper. A safe, efficient dressing peeling trajectory is first optimized in a simulated environment and then transferred into a real-world scenario.}
    \label{fig:cover}
    \vspace{-1em}
\end{figure}

Despite the void of previous work in wound redressing, previous works have shown that robots can perform autonomous safety-critical deformable manipulation tasks, such as robot-assisted dressing (cloth) \cite{li2021provably} and human arm repositioning \cite{peiros2023finding}.
Other works in surgical robotics, such as tissue grasping \cite{Tagliabue_2020}, tensioning \cite{Nguyen_2019} and dissection \cite{Shademan_2016}, also provide insight into modeling and control of deformable tissues.
However, peeling between two deformable bodies considering safety is still an unexplored area.
Authors in \cite{9561030} study peeling off a velcro tape from a rigid surface. Despite similarities, their approach does not consider a multi-softbody system and ignores safety criteria. 

In this paper, we focus on the task of wound dressing removal. Our technique involves designing a new adhesion simulation to extend our previous works on tissue modeling using position-based dynamics \cite{Fei_2021, Xiao_2024, Nikhil_2024}.
We exploit the simulation's differentiability to optimize the detachment trajectory, incorporating a set of geometric and adhesion-aware constraints based on particle dynamics.
The resultant robot behaviors exhibit careful manipulation and retraction of the dressing to expose the wound while maintaining a low force profile to achieve minimum damage to the tissue.
Our contributions in this paper can be summarized as
% Several studies have preliminarily investigated retraction strategies through simulation \cite{Patil_2010}, \cite{Jansen_2009}. Although these works utilize physics-based simulations, such as FEM, for generating retraction trajectories, they do not account for the connection properties of tissue layers. Additionally, their scenarios are simplified and do not require spatial reasoning regarding connection damage and applied strain.
\begin{itemize}
    \item We propose a geometry-based approach for modeling the dynamics of adhesiveness between two deformables and the fracture mechanism of adhesive attachment. 
    \item We developed a Model Predictive Control (MPC) method for optimizing wound dressing removal trajectories with adhesion-aware boundary constraints, accounting for the wound's shape and controlling the potential damage around the wound.
    \item The ability of our method is shown in real-world wound care experiments with various combinations of wound shape, wound dressing type, and body parts.

    % \item We developed the quasi-static nominal dynamics (employing position-based dynamics, PBD) to generate a nominal peeling trajectory through via-points, enhanced by probabilistic via-point movement primitives.
    % \item Our approach is designed to be adaptable to varying stiffness distributions and uncertainties without relying on perfect system dynamics modeling, predefined cost functions, or any specific form of connection constraints.
    % \item We developed a controller and planning algorithm based on stochastic MPC/MPPI (Model Predictive Path Integral), leveraging previous trajectory distributions to inform decision-making.
\end{itemize}

%% file: tex_files/related_work.tex
\section{Related Works}

\subsection{Modeling of Peeling Dynamics for Manipulation}

In the literature, most prior work on the manipulation of peeling, which focuses on identifying, estimating, and controlling the physical properties of objects, can be broadly classified into two categories: Lagrangian-based mechanics, geometrical analysis, and learning-based approaches.

In the Lagrangian-based mechanics approach, some researchers apply traditional Lagrangian mechanics to model the peeling process \cite{Hao_Long_2024, Gabriele_2023}. These methods use classical mechanical and energy-based principles to describe how materials behave under different forces, with force equilibrium being computed through numerical energy evaluations. In the geometrical analysis approach, methods focus on the topological aspects of the peeling process, such as the shape and trajectory of the peeling trajectory \cite{Andrew_2024_TASE}. The core of this approach involves formulating a reduced-order model for planning. Recently, learning-based methods have gained popularity in robotic research, particularly for addressing the dynamic properties of tissues. These methods use algorithms like deep learning \cite{Hang_2021, Liwei_2022_Sensors} and reinforcement learning \cite{Chen_2024_vegetablepeelingcasestudy, Scheikl_2023, yuan2021multi} to directly learn control strategies.

Several studies have explored tissue retraction tasks in surgical robotics, which involve motions similar to peeling \cite{Scheikl_2023_RAL, Tagliabue_2021_RAL, Ameya_2021_ISMR, Daniele_2021_ISMR, Sachin_2010_ICRA}. However, detailed modeling of tissue connection and attachment has not been explicitly addressed. Surgical peeling presents an even more challenging scenario due to the complexity of handling deformable objects and varying boundary attachment stiffness. Unlike simpler retraction tasks, peeling involves interactions between tissues, where the stiffness at attachment points should be handled. 

% This variation adds layers of complexity to the task, making it difficult for current methods to effectively handle such cases. These methods either require significantly more complex setups or are limited by their inability to accurately model the diverse behaviors of various materials, which can be critical in surgical environments. As a result, there remains a gap in developing models that can robustly handle the complexities associated with surgical peeling tasks.

% peeling strategies from data. More recent approach is in robotics manipulation in paper folding \cite{yuan2021multi}.

% Although the , few researchers have attempted to introduce robotics manipulation peeling problem \cite{} into surgical peeling field.

\subsection{Differentiable Simulation for Trajectory Optimization} 
Differentiable simulators provide gradients of the output within a computational workflow. This unique feature has broad applications in robotic manipulation tasks involving deformable objects. The authors of \cite{Fei_2021, Xiao_2024, Nikhil_2024} present a differentiable simulation based on position-based dynamics (PBD), demonstrating success in soft tissue manipulation using visual feedback by bridging the sim-to-real gap. Other popular simulation frameworks include projective dynamics (PD) \cite{Tao_Du_2021}, the finite element method (FEM) \cite{Tianju_2023}, and the material point method (MPM) \cite{Heiden_2021}. These frameworks are applied in various fields, such as general particle simulation \cite{Siwei_2024}, optimized soft robot design \cite{Dubied_2022}, deformable object manipulation \cite{Dongzhe_2024_RAL}, and tactile modeling \cite{Zilin_2024difftactile}.

Robotic peeling for wound care detachment involves managing a range of stiffnesses and highly deformable tissues, which significantly complicates simulation. The variability in attachment layer stiffness further complicates the accurate replication of realistic conditions. Unfortunately, none of the differentiable simulation methods discussed have addressed detachment involving changes in stiffness.

%% file: tex_files/method.tex
\begin{algorithm}[!t]

    \caption{Wound dressing peeling algorithm}
    \label{alg:overall}
    
        \SetKwInOut{Input}{Input}
        \SetKwInOut{Output}{Output}
    
        \Input{Multi-body system state $\mathbf{x}_0 : \{\mathbf{x}^s_0, \mathbf{x}^d_0\}$, the system's connectivity graph $\graph : \{\graphs, \graphd, \graphsd\}$, constraints stiffness parameters $\mathbf{k}$}
    
            $\mathcal{T} \gets \{\}$
    
            $\vre_t \gets \text{checkRemoval}(\mathbf{k}, \graphsd)$ \algorithmiccomment{Eq.~\ref{eqn:fracture}}
    
            \While {$\vre_t \neq \graphsd$} {
    
                $V^1_t, V^2_t \gets \text{AdhesionBoundaries}(\vre_t)$
                \algorithmiccomment{Eq.~\ref{eqn:search_adhesion_boundary}.\ref{eqn:v1}.\ref{eqn:v2}}
    
                $u_t \gets \text{MPC}(V^1_t, V^2_t, u_{t-1}, \statew, \mathbf{k})$
                \algorithmiccomment{Alg.~\ref{alg:mpc}}
    
                $\mathbf{\hat x}_{t+1} \gets f(\statew, u_t^*, \mathbf{k})$
    
                $\mathbf{k} \gets \text{updateAdhesion}(\mathbf{\hat x}_{t+1}, \mathbf{k}, \graphsd)$
                \algorithmiccomment{Eq.~\ref{eqn:fracture}}
                % \tcp{React to change in adhesion}
                
                $\mathbf{x}_{t+1} \gets f(\statew, u_t^*, \mathbf{k})$
                \algorithmiccomment{react to detachment}
                
                $\vre_{t+1} \gets \text{checkRemoval}(\mathbf{k}, \graphsd)$
                \algorithmiccomment{Eq.~\ref{eqn:check removal}}
    
            }    
            \Return $\mathcal{T}$

\end{algorithm}

% \begin{figure}[tbp]
%     \centering
%     \includegraphics[width=0.95\linewidth]{images/title.png}
%     \caption{Suture thread estimation and manipulation are key subtasks in autonomous suturing. We develop a spline reconstruction and grasping method that leverages reliability information for robustness.}
%     \label{title}
% \end{figure}

\section{METHODS}
We formulate the safe wound dressing removal problem as a finite horizon trajectory optimization problem under detachment awareness:
\begin{equation}
\label{eq:dynamic_system}
\begin{split}
    \min_{\xi} &\ \sum_{t=1}^T \mathcal{E}(\mathbf{x}_t) + \mathcal{R}(\xi) \\
    \mathrm{s.t.}~ & \mathbf{x}_{t+1} = f(\mathbf{x}_{t-1}, u_t),  u_t \in \xi(t)\\
    & \graphsd(\mathbf{x}_{t+1}) = \{0, 1\}\\
    % \left\{\xi_{1}, \xi_{2}, \cdots, \xi_{t}\right\} \\
\end{split}
\end{equation}
where $\xi$ is a trajectory of robot end-effector position $u_t$, $f$ is a function that describes the dynamics of the multi-body system consisting of a wound dressing and human skin. $\mathcal{E}$ and $\mathcal{R}$ are safe peeling cost functions and regulation of the end-effector's trajectory respectively. $\graphsd$ is the connectivity between the wound dressing and human skin due to adhesion, while $0$ is detached and $1$ is attached. Later in this section, we first discuss our proposed modeling approach for the multi-body system in \autoref{sec:peeling_model}, followed by the definition of our design of cost function $\mathcal{E}$ to achieve safe wound dressing peeling in \autoref{sec:peeling_and_safety_objective}. Lastly, we discuss how we solve the problem via MPC optimization in \autoref{sec:MPC}. Our overall algorithm is shown in \autoref{alg:overall}.
% We say the control trajectory $\xi(\tau)$ can achieve the dressing removal goal enforced as a constraint $\mathbf{C}_\text{remove}$, and can maintain safety by minimizing the time integral of a safety-related cost function $\mathcal{E}$ under system dynamics $f$.
% \begin{itemize}
%     \item In our paper, two critical concepts are nominal modeling and stochastic Model Predictive Control (MPC)/Model Predictive Path Integral (MPPI).
%     \item Our method is composed of three primary steps. First, we employ Position-Based Dynamics (PBD) as the nominal model to simulate the surgical application scenario. Second, leveraging this nominal model, we generate a nominal peeling trajectory using via-points and probabilistic via-point movement primitives. Finally, we develop a Model Predictive Control (MPC) or Model Predictive Path Integral (MPPI) controller based on the previously generated trajectory distribution.
% \end{itemize}
% \begin{algorithm}[!t]
%     \caption{Residual Mapping and Online Stiffness Optimization in a PBD simulation}
% \end{algorithm}
\subsection{Peeling Dynamics Modeling}\label{sec:peeling_model}
\textbf{Softbody Modeling:} We aim at modeling the behavior of a multi-soft body system with state $\statew$ whose element $\mathbf{x}^i_t \in \mathbb{R}^3$ represents a particle position. 
The multi-body system contains a region of human skin and the wound dressing. Their states are represented by $\states$ and $\stated$ respectively.
As the peeling task involves a variety of different object properties, we use the Extended Position-Based Dynamics (XPBD) modeling approach for efficient and stable simulations. Our approach incorporated a set of distance constraints $\mathcal{C}$ directly into the position update of $\Delta\mathbf{x}$. 
% By adjusting stiffness parameters $\mathbf{K}$ linked to the constraints, XPBD provides flexibility in achieving realistic behaviors for a wide range of materials for peeling tasks. 
We can model the XPBD dynamic with energy minimization as
\begin{equation}
\label{eq:pbd_sim}
\begin{split}
\mathbf{x}_{t+1} &= \argmin_{\mathbf{x}_{t+1}} \sum_{i, j \in \graph} E_{t+1}^{i, j}\\
& = \argmin_{\mathbf{x}_{t+1}} \sum_{i, j \in \graph} \frac{1}{2} k^{i,j} \mathcal{C}(\mathbf{x}^i_t, \mathbf{x}^j_t)^2,\\
s.t.&\ \mathbf{x}^u_t = u_t, \mathbf{x}^u_t \in \stated \\
& \{\mathcal{G}^s, \mathcal{G}^d, \graphsd \} \subseteq \mathcal{G}
\end{split}
\end{equation}
in which $\graph$ is the connectivity graph of the whole multi-body system containing dressing and skin internal connectivity $\graphd, \graphs$, and inter-connectivity $\graphsd$. $\mathbf{x}^u_t$ is a particle on the wound dressing that is rigidly coupled to a robot's end-effector position $u_t$. A distance constraint is formulated as a linear spring constraint $\mathcal{C}(\mathbf{x}^i, \mathbf{x}^j) = \|\mathbf{x}^i-\mathbf{x}^j\| - \|\mathbf{x}^i_0-\mathbf{x}^j_0\|$. The stiffness value $k^{i,j}$ is defined for individual constraints, enabling potentially different strengths of constraint satisfaction. We use distance constraints to model both volumetric human skin and thin-shell wound dressing over $\graphd, \graphs$.
% Note that we use only distance constraints for favor of their simplicity. More sophisticated constraints emcompassing aspects such as preservation of volume and shape can be flexibly incorporated as well.

\begin{table}[t]
\setlength\tabcolsep{0.3em}
\centering
\caption{Important symbols}
\begin{adjustbox}{width=0.48\textwidth}
    \begin{tabular}{cl}
    \toprule
Symbols & Meanings\\ 
  \midrule
$\statew$ & particles' state of entire multi-softbody system \\
$\states \subset \statew$ & particles' state of the human skin \\
$\stated \subset \statew$ & particles' state of the wound dressing\\
$\statedad \subset \stated$ & particles' state of adhesive region on the wound dressing\\
$x^{i_s}$ & the $i$-th particle in $\states$\\
$x^{j_d}$ &  the $j$-th particle in $\stated$\\
$\graph$ & Graph of particle connectivity of the system\\
$\graphs, \graphd, \graphsd$ & Connectivity graph of internal of skin, dressing, \\
&and inter-connection between them\\
$\vre_t$ &  Current set of adhesion connections that have been detached\\
% &been \feimodify{detached}\\
$V^1_t, V^2_t$ & the 1st and 2nd detachment-aware adhesion boundary layers \\
    \bottomrule
\end{tabular}
\end{adjustbox}
\label{tbl:accuracy_simulation}
\end{table}

\textbf{Adhesion and Peeling:} To model adhesion between the skin and dressing for the (de-)attachment representation, we apply the spring boundary constraint from our previous work \cite{Nikhil_2024}. It assumes there exist virtual springs with zero resting length between the human skin surface and the adhesive part of wound dressing $\statedad \subset \stated$.  To avoid complication, we further assume that for any particle $\bar x^{i_d} \in \mathbf{\bar x}^d_0$, there always exists a duplicated particle $\mathbf{\bar x}^{j_s}\in \mathbf{x}^s_0$
% $x^{j_s}\in \mathbf{x}^s_0$
that has exactly the same position at initialization. We define every pair of virtual spring connections with $\graphsd \subset \graph$. The skin-dressing connectivity should only be broken for $\graphsd=\{0,1\}$ to represent both the detachment and attachment of adhesion.

Peeling off the dressing indicates detachment of adhesion between the skin and the dressing, i.e., $\graphsd: 1 \mapsto 0$.
% We propose a fracture machnism for the spring boundary constraint to simulate the peeling effect. 
After every simulation step Eqn.~\ref{eq:pbd_sim}, we decide if a boundary constraint should be fractured by checking whether its potential energy exceeds a threshold $\varepsilon$, and modifying the corresponding stiffness value to be
\begin{equation}\label{eqn:fracture}
    \begin{split}
        k^{i, j} = \begin{cases}
            0&E_{t+1}^{i, j} \ge \varepsilon, \\k^{i, j}&E_{t+1}^{i, j} < \varepsilon.
        \end{cases},\ (i, j) \in \graphsd
    \end{split}
\end{equation}
With this, adhesion detachment is achieved because those spring boundary constraints with zero stiffness value will not be satisfied. This mechanism is illustrated in Fig.~\ref{fig:boundary_and_fracture} left.

% With \feimodify{$K = \{k^{i,j}|(i, j) \in \graphsd\}$ and $\graphsd=\{0,1\}$}, 
Defining $k^{i, j} = 0$ when $\graphsd: 1 \mapsto 0$, 
we can identify which pair of the adhesion constraint is detached by the following mechanism,
\begin{equation}\label{eqn:check removal}
    \vre \doteq \{\left(i_{\text{det}}^d, j_{\text{det}}^s \right)| k^{i, j} = 0, (i, j) \in \graphsd\}
\end{equation}

\subsection{Peeling and Safety Objectives}\label{sec:peeling_and_safety_objective}
% \textcolor{red}{One paragraph clarifying why we choose some heuristic instead of global optimization}
% Our peeling goal is to achieve removal of all adhesion while regulate the induces skin deformation.
We observe that human care-givers who perform the peeling task follow a heuristic by peeling back along the skin surface. 
As a result, they can focus only on the boundary layer between detached adhesion and the remaining adhesion region, minimizing the overall skin deformation.
% As a result, only the adhesion near the boundary between the non-adhesive and adhesive regions detaches, minimizing the overall skin deformation. 
Here, we refer to this boundary as the \textbf{adhesion-aware boundary}, \textbf{adhesion boundary} in short. Inspired by that, we propose a new peeling heuristic that prioritizes extending the adhesion connection along the boundary, layer by layer, following the peeling trajectory, while simultaneously keeping the remaining attached connections in their rest positions as much as possible for safety. To achieve it, we need to identify the current \textbf{adhesion boundary}.
\begin{figure}[t]
    \centering
    \includegraphics[width=0.49\textwidth]{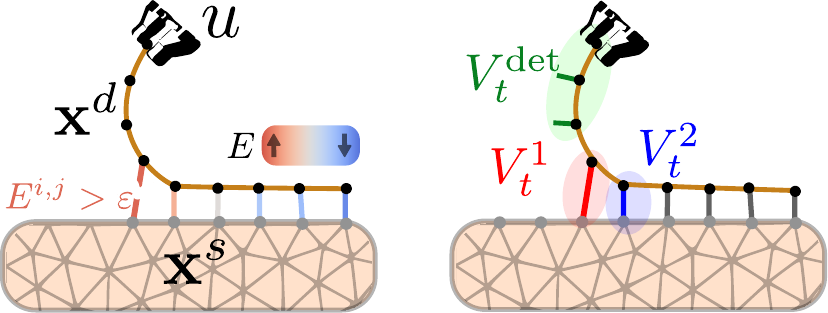}\vspace{-1.5em}
    \caption{\textbf{Left}: adhesion modeling and fracture mechanism of adhesive connections. Adhesion is modeled with distance constraints between elements in $\states$ and $\stated$. An adhesion constraint is fractured when its potential energy $E^{i, j}$ exceeds a limit $\varepsilon$. \textbf{Right}: visualization of current removed connections $\vre_t$, the first immediate adhesion layer $V^1_t$, and the second immediate adhesion layer $V^2_t$.}
    \label{fig:boundary_and_fracture}\vspace{-1.5em}
\end{figure}

\begin{figure}[t]
    \centering
    \includegraphics[width=0.5\textwidth]{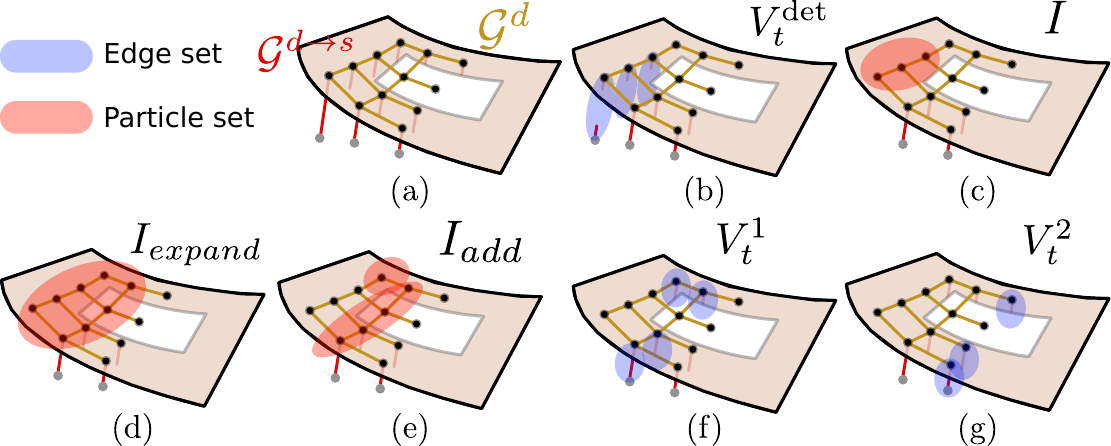}\vspace{-1.5em}
    \caption{A visualization of our procedures for finding the adhesion boundary. From figure (a)-(f), finds the 1st immediate adhesion boundary layer $V^1_t = \mathcal{D}(\vre_t)$. To further obtain the 2nd adhesion boundary layer, we apply the procedure recursively to get $V^2_t=\mathcal{D}(\vre_t \cup V_t^1)$.}
    \label{fig:adhesion_boundary}\vspace{-1.5em}
\end{figure}

We define $\mathcal{D}(\vre_t)$ as a function that tries to find the current adhesion boundary. The overall procedures of finding the adhesion boundary are followed by,
\begin{itemize}
    \item extracting a set of particles $I$ in $\stated$ whose associated adhesion have been detached
    % \[I = \{\mathbf{\bar{x}}_{det}^d\ | (\mathbf{\bar{x}}^d_i, \mathbf{\bar{x}}^s_j) \in V\}\]
    \item expanding this set $I_{\text{expand}}$ by one step (traverse ONE edge) based on the dressing's connectivity
    % \[I_{expend} = \{l_d | (l_d, i_d) \in \graphd, i_d \in I\}\]
    \item subtracting $I$ to exclude removed adhesion connections
    \begin{equation*}
        I_{\text{add}} = I_{\text{expand}} \ \backslash\ I
    \end{equation*}
\end{itemize}
Then, the current adhesion boundary is defined as 
\begin{equation}\label{eqn:search_adhesion_boundary}
    \begin{split}
         \mathcal{D}(\vre_t) & \doteq \{(i_{\text{adh}}^d, j_{\text{adh}}^s) | (i^d, j^s) \in \graphsd, i^d \in I_{\text{add}}\}.
    \end{split}
\end{equation}
This suggests that the adhesion boundary is determined through both the connectivity graph and the layer-by-layer expansion/adding. This procedures are also illustrated in \autoref{fig:adhesion_boundary}. Let the 1st immediate adhesion boundary be
\begin{equation}\label{eqn:v1}
    \begin{split}
         V^1_t = \mathcal{D}(\vre_t)
    \end{split}
\end{equation}
We would like to apply robot controls to cause fracture to $V^1_t$. However, at the same time, we want to regulate other remaining layers of adhesion boundary to stretch less. To achieve it, we further find the 2nd layer of adhesion boundary $V^2_t$ via by modifying \autoref{eqn:search_adhesion_boundary} to 
\begin{equation}\label{eqn:v2}
    \begin{split}
        V^2_t = \mathcal{D}(\vre_t \cup V_t^1)
    \end{split}
\end{equation}
Intuitively, $V^2_t$ will become the next adhesion boundary if all $V^1_t$ is successfully fractured.
% This operation finds vertices on the dressing whose adhesion hasn't been removed and that are neighboring to elements in $V$.
% Let $\vre_t$ contains elements in $\graphsd$ that has been removed by the fracture mechanism in Eqn.~\ref{eqn:fracture}.
% At every simulation step, we try to find two set $V^1_t, V^2_t$ that determine the current adhesion boundary:
% \begin{equation}
%     \begin{split}
%         V_t^1 &= D(\vre_t),\\
%         V_t^2 &= D(\vre_t \cup V_t^1).
%     \end{split}
% \end{equation}
% Intuitively, $V^1_t$ describes a layer of adhesive particles that are right next to the current peeling boundary, and $V^2_t$ is the next immediate layer of adhesive particles if $V^1_t$ has been removed.

We design an objective to guarantee peeling progress by stretching the adhesive boundary $V^1_t$, but regulate the length of adhesive connections associated with $V^2_t$. We decide to regulate only $V^2_t$ but on all other remaining connections because $V^2_t$ approximates is a good approximation to connections that are further from the current boundary, as shown in \autoref{fig:boundary_and_fracture} right. Mathematically, the objective we want to minimize is 
\begin{equation}\label{eqn:objective}
    \begin{split}
         \mathcal{H}(\statew)= & - \frac{1}{\texttt{card}(V^1_t)}\sum_{(d_1, s_1) \in V^1_t} \mathcal{C}(\mathbf{x}_{\text{det}}^{d_1}, \mathbf{x}_{\text{det}}^{s_1})\\
         & + \gamma \cdot\frac{1}{\texttt{card}(V^2_t)}\sum_{(d_2, s_2)\in V^2_t} \mathcal{C}(\mathbf{x}_{\text{det}}^{d_2}, \mathbf{x}_{\text{det}}^{s_2}),\\
    \end{split}
\end{equation}
where $\gamma$ is a parameter that we tune, and \texttt{card} represents cardinality. 
This objective aligns well with gradient-based optimization methods because the differentiable nature of constraint-based simulation allows for efficient gradient computation across all distant constraints. The new heuristic leads to human-like safe action and can adapt to different scenarios with optimization.

% A deformable object here can be define as a set of vertexes $V = \{v_1, v_2 ... v_n\}$ and a set pf triangles $T = \{t_1, t_2 ... t_m\}$ where $t_i = \{(v_a, v_b, v_c)|v_a, v_b, v_c \in V\}$. Each edage $E = \{(v_i, v_j)|v_i, v_j \in t_n, t_n \in T\}$ with the vertexes $V$ we define above forms a connected undirected graph $G = (V, E)$.
% \\
% For the attachment layer, we define it as a thinshell of a set of vertexes $V_{attach} = \{v_i|v_i \in V\}$ and a set of triangles $T_{attach} = \{t_1, t_2 ... t_n\}$ where $t_i = \{(v_a, v_b, v_c)|v_a, v_b, v_c \in V_{attach}\}$. Edges $E_{attach} = \{(v_i, v_j)|v_i, v_j \in t_n, t_n \in T_{attach}\}$. Same as the deformable object, we can define a connected undirected graph $G_{attach} = (V_{attach}, E_{attach})$.
% \\
% In our method, we first define a control vertex $V_{control} \in V_{attach}$ which donate the control point we will manipulate in the process. Our method first find the area $V_{dettached} \in V_{attach}$ which the vertexes $V_{dettached}$ are the vertexes that have been dettached during the process. Then we start from $V_{control}$ to perform a graph searching algorithm to find the neighbor vertex $\{V_1 = \{ v_i|\{v_i, v_j\} \in E, v_i \notin V_{dettached}, v_j \in V_{dettached} \}\}$ and vertex $\{V_2 = \{ v_i|\{v_i, v_j\} \in E, v_i \notin V_1, v_j \in V_1 \}\}$ that are neighbors to $V_1$. The graph searching algorithm is showed on \{to do\}. Our goal is focus on the optimization on these two point sets $V_1, V_2$ which will be explained on next section.

\begin{algorithm}[!t]
    \caption{Model predictive end-effector control}
    \label{alg:mpc}
    \SetKwInOut{Input}{Input}
    \SetKwInOut{Output}{Output}

    \Input{Adhesion layers $V_t^1, V_t^2$, end-effector position $u_t$, system state $\statew$, stiffness $\mathbf{k}$}

       $\mathbf{A} \gets \{A_1,A_2, \dots \}$
       \algorithmiccomment{Sample action seeds}

       \For {numIterations} {
            $\mathbf{X}_{t+h} \gets \text{parallelSimulationSteps}(\statew, u_t, \mathbf{A}, h)$
            
            $\mathbf{L} \gets \text{computeLoss}(\mathbf{X}_{t+h}, V_t^1, V_t^2, \mathbf{k})$
            \algorithmiccomment{Eq.~\ref{eqn:mpc_loss}}

            $\mathbf{A}.\vec{v} \gets \mathbf{A}.\vec{v}- \lambda \frac{\partial L}{\mathbf{A}.\vec{v}} $
            \algorithmiccomment{Gradient Descend}
       }

    $i^* \gets  \argmin_i L_i, L_i \in \mathbf{L}$
    \algorithmiccomment{Pick the best}

    $A^* \gets A_{i^*}$

    $u_{t+1} \gets u_{t} + A^*.\Vec{v} \cdot A^*.s$

    \Return $u_{t+1}$

\end{algorithm}
\subsection{Model Predictive Control}\label{sec:MPC}
Wound dressing peeling will be formulated as an optimization problem where MPC is used to generate a trajectory for optimal wound dressing peeling. Let each control action to the MPC be $A: \{\Vec{v}, s\}$ where $\Vec{v}$ is a vector that represents a control direction and $s$ represents a control step size. We tune $s$ as a hyper-parameter and optimize only $\vec{v}$. The MPC will recursively apply the same action to predict the system behavior for a short horizon $h$. In other words, we assume that the MPC always predicts with linear trajectory.

Besides using Eqn.~\ref{eqn:objective} as our loss function to minimize, we want to constraint the robot end-effector to avoid penetration to human skins. For efficient gradient-based optimization, we use a penalty-based method with penalty function $\mathcal{P}$ to achieve the non-penetration constraint:
\begin{equation}
    \begin{split}
        \mathcal{P}(u_t) = \exp\Bigl(
        -\min\bigl(
        \phi_{\mathbf{x}_0^s
        }(u_t)-\sigma, 0\bigr)
        \Bigr),
    \end{split}
\end{equation}
where $\phi_{\mathbf{x}_0^s} : \mathbb{R}^3 \rightarrow \mathbb{R}$ represents a signed distance function (SDF) of the geometry represented by $\mathbf{x}_0^s$. The SDF can be analytically obtained for simple geometry and estimated by methods such as \cite{zhi2022diffco} in more complex scenarios. 

We also apply a smoothness penalty loss, $\mathcal{S}$, that regularizes the change in action direction w.r.t to the previous action direction:
% We also wound like to improve the smoothness of the end-effector's trajectory, by applying a smoothness penalty loss $\mathcal{S}$ that regularizes the change in action direction w.r.t to the previous action direction:
\begin{equation}\label{eqn:smooth function}
    \begin{split}
        \mathcal{S}(A_{t}, A_{t-1}) = ||A_{t}.\vec{v} - A_{t-1}.\vec{v}||
    \end{split}
\end{equation}

Lastly, the MPC seeks to minimize the loss function
\begin{equation}\label{eqn:mpc_loss}
    \begin{split}
        L = \mathcal{H}(\mathbf{x}_{t+h}) + \alpha\sum_{t=t_0}^{t+h}\mathcal{P}(u_{t}) - \beta \cdot \mathcal{S}(A_t, A_{t-1}).     
    \end{split}
\end{equation}
% \textcolor{red}{add differentiable collision avoidance.}
We sample many seeds for different actions, predicting and optimizing them in parallel. The optimal action $A^*$, is selected from all seeds after several steps of gradient-based optimization, is added to update the end-effector position. The MPC optimization is detailed in Alg.~\ref{alg:mpc}.

%% file: tex_files/experiments.tex
\section{Experiment}
We validate our method in several real-world scenarios. These experiments are conducted using the Franka Emika Panda Robot to peel off real wound dressing from foam phantoms. The experiments are first carried out multiple times in a controlled scenario for quantitative analysis, then we show a demo experiment in which our algorithm performs wound dressing peeling on a real human subject.

\textbf{Experiments Setup:} The controlled scenarios are set up following Fig.~\ref{fig:phantom_setup}.
A thin-shell 7x7 inches squared foam phantom with 0.1 inches thickness, is hung and tensioned by 3d printed mounts at every corner to simulate skin.
During each experiment, a wound dressing is adhered to the top side of the phantom, and a Panda Robot will attempt to peel the dressing off the phantom. 
We test our method with two types of wound dressings that have different dimensions. They are WD-1 with size of 4x4 inches and WD-2 with size of 4.5x3.5 inches. 
We place an iPad at the bottom of this physical setup, facing upward. The iPad's front camera is able to capture real-time RGBD video of the phantom for us to evaluate the deformation that the phantom undergoes during experiments. Additionally, we labeled 6x8 green landmarks at the back phantom. Later, we use a visual feature tracking method CoTracker \cite{karaev2023cotracker} to track landmarks in RGB videos. Combining with depth information, we can obtain deformation estimation in 3D. In the experiments where real humans are involved, we wrap a foam sheet around the subject's leg and adhere a wound dressing onto the foam.

\begin{figure}[t]
    \centering
    \includegraphics[width=0.5\textwidth]{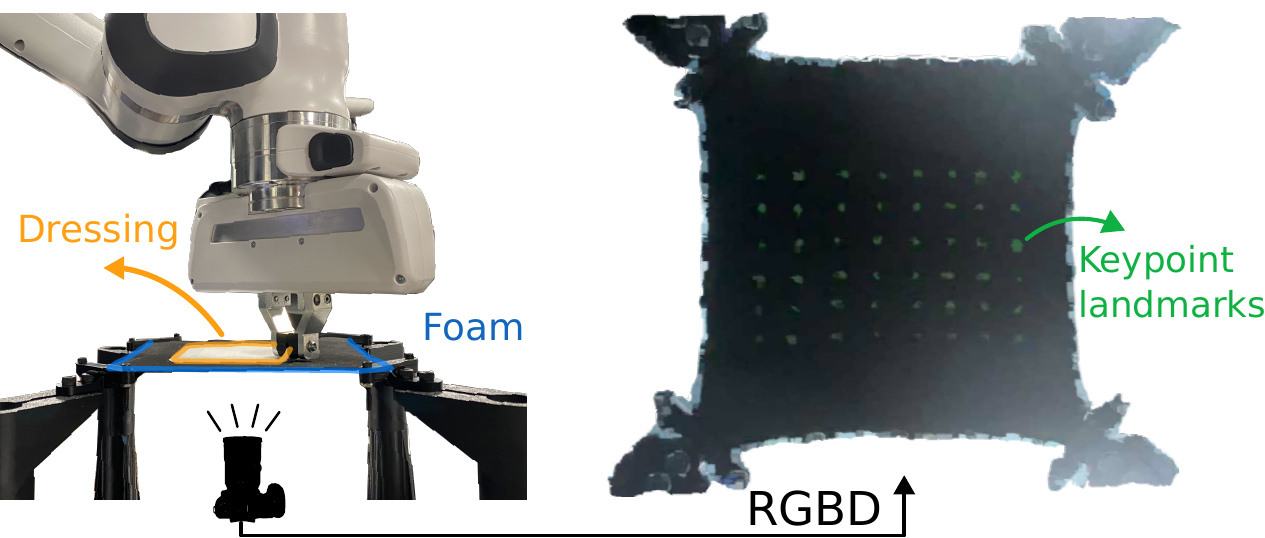} \vspace{-1.5em}
    \caption{Our physical setup where the controlled experiments' results were collected. The wound dressing (in orange) is adhered to a foam phantom (in blue). The RGBD camera sees the bottom side of the phantom. There are 6x8 green landmarks labeled on the back of the foam. We evaluate the deformation of the phantom by combining depth and landmark tracking from the RGB image. An example point cloud output from the camera is shown on the right.}
    \label{fig:phantom_setup}
    \vspace{-1em}
\end{figure}
% \begin{figure*}[t]
%         \centering
%         \includegraphics[width=0.8\textwidth]{images/simulation_exp.png}
%         \caption{Simulation on different Setups}
%         \label{fig:simulation exp}
%     \end{figure*}

    % \begin{figure}[t]
    %     \centering
    %     \includegraphics[width=0.5\textwidth]{images/phantom_comparison.pdf}
    %     \caption{Comparison of different wound peeling method on foam phantom.}
    %     \label{fig:phantom_experiment}
    % \end{figure}
    %     \begin{figure}[t]
    %     \centering
    %     \includegraphics[width=0.5\textwidth]{images/phantom_plot.pdf}
    %     \caption{Plot of maximum vertical and planar motion caused by 3 comparison methods on 4 different peeling scenarios.}
    %     \label{fig:phantom_plot}
    % \end{figure}

\textbf{Implementation Details:}
The proposed method is implemented in Python and tested on a machine with NVIDIA RTX A6000 GPU. The PBD simulation is implemented with PyTorch.
In all experiments, we start peeling after having the robot grasp at a corner of the wound dressing.
The end effector's z-direction points downward and the y-direction points at the center of the wound dressing.
Peeling trajectories are generated by first optimizing in a simulated environment and then are transferred to real scenarios. We manually define the end-effector's orientation such that its z-direction is always looking at the point $u_0 - [0, 0, 0.1m]$ and its x-direction stays constant. For the simulation, we implement a Model Predictive Controller using the Eqn.~\ref{eqn:mpc_loss} with tuned parameter $\alpha, \beta$ as the loss function with tuned parameter $\gamma$ we mentioned in Eqn.~\ref{eqn:objective}. In the phantom experiment, we use 
parameters $\alpha=0.1, \beta=0.01, \gamma=1$. In the real human experiment, we increase the $\gamma=1.5$ as for more conservative and safer peeling behavior. Our MPC step uses a horizon $h=10$, 60 seeds based on Gaussian sampling on $X, Y, Z$ axis independently with $\sigma = 0.05$ and $(\mu_x, \mu_y, \mu_z) = u_t$, and optimizes for 1 step. We increase the optimization step for the real human experiment to 2 steps for better trajectory quality.
% The optimization of flat skin case takes 10 minutes, and 23 minutes for the cylindrical scenario.

\begin{table}[t]
\setlength\tabcolsep{1.8em}
\centering
\caption{Comparison of maximum landmarks displacement $D_{max}$ and average landmarks displacement $D_{mean}$ from different methods. Our method achieves significantly smaller displacement than other methods in both metrics. $n$ is the number of repetitions in each controlled scenario}
% \caption*{\footnotesize Smaller note of table that describes what the table is all about.}
\begin{adjustbox}{width=0.48\textwidth}
    \begin{tabular}{l|cc}
    \toprule
         & $D_{max}$ (mm) & $D_{mean}$ (mm)\\ 
  \midrule
Up-WD1 (n=5) & $43.5\pm9.9$ & $12.8 \pm 3.1$\\
Arc-WD1 (n=5) & $29.3\pm1.7$ & $9.7\pm0.6$\\
Ours-WD1 (n=20) & $\mathbf{20.6\pm4.0}$ & $\mathbf{4.7\pm 0.9}$\\
\midrule
Up-WD2 (n=5) & $47.2\pm2.6$ & $18.7\pm4.1$ \\
Arc-WD2 (n=5) & $42.6\pm1.8$ & $17.9\pm1.7$ \\
Ours-WD2 (n=20) & $\mathbf{27.2\pm2.6}$ & $\mathbf{5.5\pm0.8}$ \\
\bottomrule
\end{tabular}
\end{adjustbox}
\label{tbl:phantom_experiments}
\end{table}

\begin{figure}[!t]
    \centering
    \includegraphics[width=0.48\textwidth]{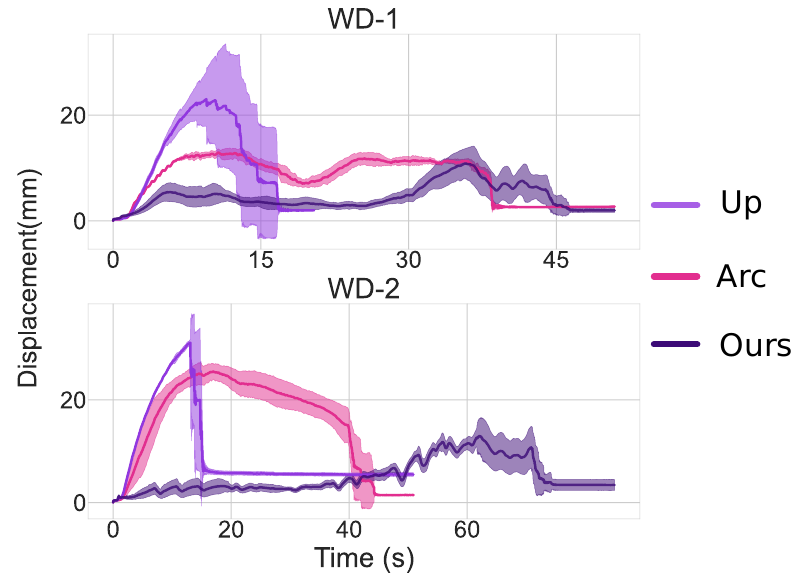}
    \vspace{-2em}
    \caption{Comparison of mean and standard deviation of landmarks displacement results overtime of different methods. The plot has shown that our method causes the lowest average landmark displacement overall, showing that the method can achieve the peeling goal in a safer manner.}
    \label{fig:phantom_plot}
    \vspace{-1.5em}
\end{figure}

% \begin{figure}[!t]
%     \centering
%     \includegraphics[width=0.5\textwidth]{images/distribution.png}
%     \caption{Distribution plot of $D_{mean}$ over different method. This plot has shown that over 5 repetition for baseline methods and 20 repetition for our method, the distribution is within acceptable range and shows the robustness of our result.}
%     \label{fig:distribution plot}
%     \vspace{1.em}
%     % \vspace{1em}
% \end{figure}

\begin{figure*}[t]
    \centering
    \includegraphics[width=0.98\textwidth]{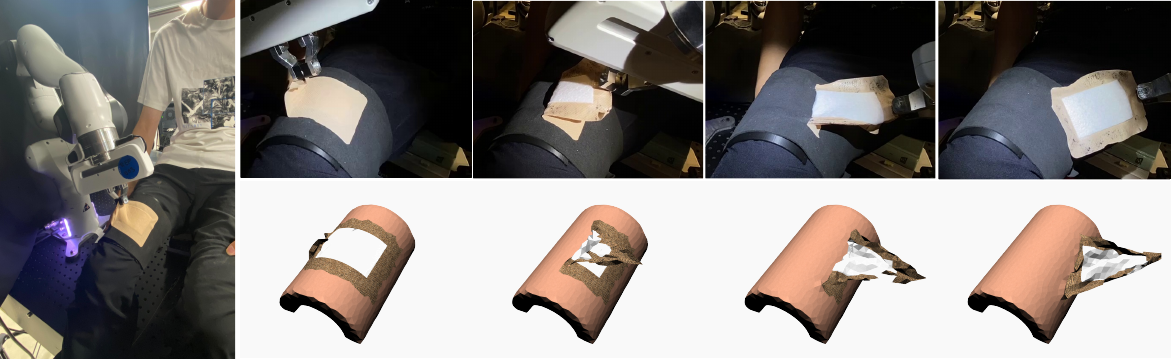}\vspace{-0.2em}
    \caption{Results of transferring an optimized trajectory from simulation on to peeling a wound dressing (WD-2) from a human subject's leg. Following this trajectory, the robot's end effector carefully peels the dressing off while complying with the leg's geometry. It achieves full removal of the wound dressing successfully, showing our method's potential application in complex real-world scenarios.}
    \label{fig:real_exp_on_human}\vspace{-1em}
\end{figure*}
\begin{figure}[!t]
    \centering
    \includegraphics[width=0.5\textwidth]{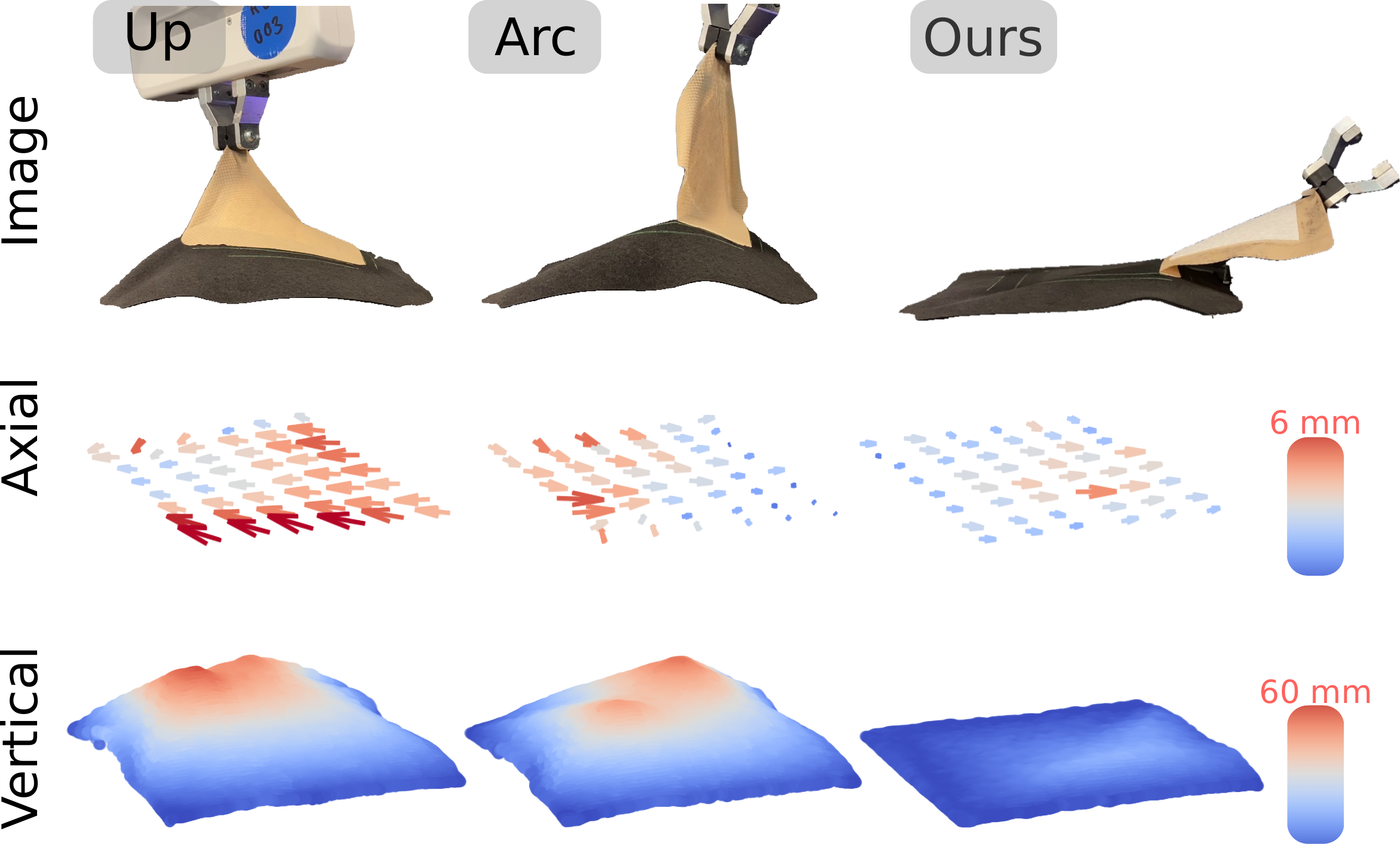}\vspace{-1.2em}
    \caption{Visual comparison of maximum magnitudes of skin deformation during peeling with different methods. Based on results from Fig.~\ref{fig:phantom_plot}, we select time points when each method causes its maximum deformation to visualize. The Up and Arc baseline methods induce significantly more axial and vertical deformation at the beginning. Deformation from our method increases towards the end but is still smaller in comparison. }
    \label{fig:phantom_deformation_vis}
    \vspace{-1.5em}
\end{figure}
\textbf{Experiment results}:
We aim to validate our method for $n$ number of repetitions in each controlled scenario where $n$=20. 
Therefore, we test on 2 wound dressings with in total of 40 repetitions. The proposed method is compared to 2 heuristic-based baseline methods. 
They are 1.\textbf{Up}: slowly move the end-effector vertically up, and 2. \textbf{Arc}: move in a circular path whose center is in the direction of the dressing center.
We set $n$=5 for every baseline method.
Displacement of the landmarks is used as our main metric for evaluation. let the magnitude of displacement of the $i$-th landmark at time point $t$ be denoted as $\|d^i_t\|$, $t \in [0, T], i \in [1, N]$. We use the following two metrics for evaluation:
\begin{itemize}
    \item $D_{max}$ (mm): Maximum displacement of all landmarks during a peeling trajectory:
    \begin{equation*}\label{eq:evaluation_matrics1}
        \begin{split}
        D_{max} = \max \|d_t^i\|. \\
        \end{split}
    \end{equation*}
    \item $D_{mean}$ (mm): displacement averaged across landmarks:
    \begin{equation*}\label{eq:evaluation_matrics2}
    \begin{split}
        D_{mean} = \frac{1}{nT}\sum_i^n \sum_t^T \|d_t^i\|. \\
        % t \in [0, T], i \in [1, n]
    \end{split}
    \end{equation*}
    % Where $d_{t}^{i}$ is the deformation of $i$th landmark at time $t$, $T$ is the duration time of the experiment and $n$ is the number of landmarks we labeled in the experiment.
\end{itemize}
These metrics are chosen to reflect the maximum potential discomfort and overall discomfort to the skin tissue that can be caused by different methods.
% \begin{itemize}
%     \item $D_{max}$ (mm): average of the displacement of all landmarks to measure planar deformation.
%     Average of the maximum deformation of the landmark during each set of experiment.
%     \item $D_{average}$ (mm): average change in depth of the phantom region to measure verticle deformation.
% \end{itemize}

Our overall results are summarized in Table~\ref{tbl:phantom_experiments}. It shows that in comparison to baseline methods, our method achieves the lowest average $D_{max}$ and $D_{mean}$ within two wound dressing scenarios.
Fig.~\ref{fig:phantom_plot} shows the distribution of landmark displacements overtime for different methods. The average landmark displacements of the baseline methods increase rapidly at the beginning of the phantom experiment and maintain at a high level until the wound dressing is fully detached from the foam. This is because their initial actions are largely orthogonal to the adhesion surface, and thus more force is required to detach adhesion for a larger area. On the other hand, our method maintains $D_{mean}$ at a significantly lower level as it is informed by the designed heuristic to break only the immediate adhesion connection will minimize deformation in other areas. 

To better show this, we visualize the maximum deformation that each phantom undergoes in Fig.~\ref{fig:phantom_deformation_vis}. Because our landmark tracking is sparse, we visualize the axial displacement on the xy-plane and vertical deformation separately. Axial displacements are shown from landmarks tracking and vertical deformation is obtained from depth sensing. The results show our method's maximum deformation. Both baseline methods induce more than 5mm axial maximum displacement and 50mm maximum vertical displacement, whereas for ours the numbers are around 4 mm and 20 mm. Our results show that peeling along the skin surface, our MPC loss should reduce damage to the skin tissue since we measured less displacement.
% , and that with designed heuristic, our method can achieve similar peeling behavior while generalize to different wound dressing shapes. 

We further show its ability to work on another skin configuration in a demo experiment where a real human subject is involved, as shown in Fig.~\ref{fig:real_exp_on_human}. We build a simulated environment where we simulate skin on human legs with a cylindrical shell. The optimized simulated trajectory peels the wound dressing closely along the shell surface. After transferring the trajectory, we demonstrate a peeling procedure can potentially apply to a real human subject.

%% file: tex_files/conclusion.tex
\section{Discussion \& Conclusion}

In this paper, we proposed an optimization-based method for generating safe wound dressing peeling trajectory, the first step towards automated redressing for wound care. We proposed a position-based formulation to model the adhesion connectivity within a multi-softbody system. We developed a detachment mechanism using energy formulation of constraints to ensure safe trajectory generation. Therefore, the adhesion connection can be stretched safely within a specific range to accommodate peeling dynamics.
% where the adhesion connection are stretched allow as to further simulate the dressing peeling dynamics. 
With the model, we proposed an MPC controller for safe removal of wound dressing that is able to achieve peeling behavior similar to clinical care-givers. We validate our method's efficacy in both skin phantom experiments and a real-world human demo experiment.
One limitation of our current method is the real-to-sim gap. This can be addressed through system identification or by using real-world sensor feedback, like RGB-D cameras, to adaptively adjust the trajectory. Optimizing adhesion-aware trajectory with a closed-loop system can reduce pain during dressing removal, improving patients' quality of life. Future work will involve clinical experiments to confirm that minimizing deformation correlates with less pain, paving the way for more patient-friendly wound care.